\documentclass[10pt,twocolumn,letterpaper]{article}

\usepackage[pagenumbers]{cvpr} 

\usepackage{graphicx}
\usepackage{amsmath}
\usepackage{amssymb}
\usepackage{booktabs}

\usepackage[usenames,dvipsnames]{xcolor}
\usepackage{colortbl}
\usepackage{paralist}
\usepackage[utf8]{inputenc}
\usepackage{soul}
\usepackage{multirow}
\usepackage{float}
\usepackage{cuted}
\usepackage{capt-of}
\usepackage{comment}
\usepackage{algorithm}
\usepackage{algpseudocode}
\usepackage{siunitx}

\newif\ifshowcomments
\showcommentstrue

\newcommand{\figref}[1]{Fig.~\ref{#1}}
\newcommand{\tabref}[1]{Tab.~\ref{#1}}
\newcommand{\secref}[1]{Sec.~\ref{#1}}

\newcommand{\equref}[1]{Eq.~(\ref{#1})}

\newcommand{\datasetname}{Hi4D\xspace}

\usepackage[pagebackref,breaklinks,colorlinks]{hyperref}

\usepackage[accsupp]{axessibility}  

\usepackage[capitalize]{cleveref}
\crefname{section}{Sec.}{Secs.}
\Crefname{section}{Section}{Sections}
\Crefname{table}{Table}{Tables}
\crefname{table}{Tab.}{Tabs.}

\def\cvprPaperID{1057} 
\def\confName{CVPR}
\def\confYear{2023}

\begin{document}

\title{\datasetname: 4D Instance Segmentation of Close Human Interaction}

\author{Yifei Yin \quad Chen Guo \quad Manuel Kaufmann \quad Juan Jose Zarate \quad Jie Song \quad Otmar Hilliges
\\ETH Zurich \vspace{-6 mm}
}
\maketitle

\vspace{-1em}
\begin{strip}
\centering
\includegraphics[width=\textwidth]{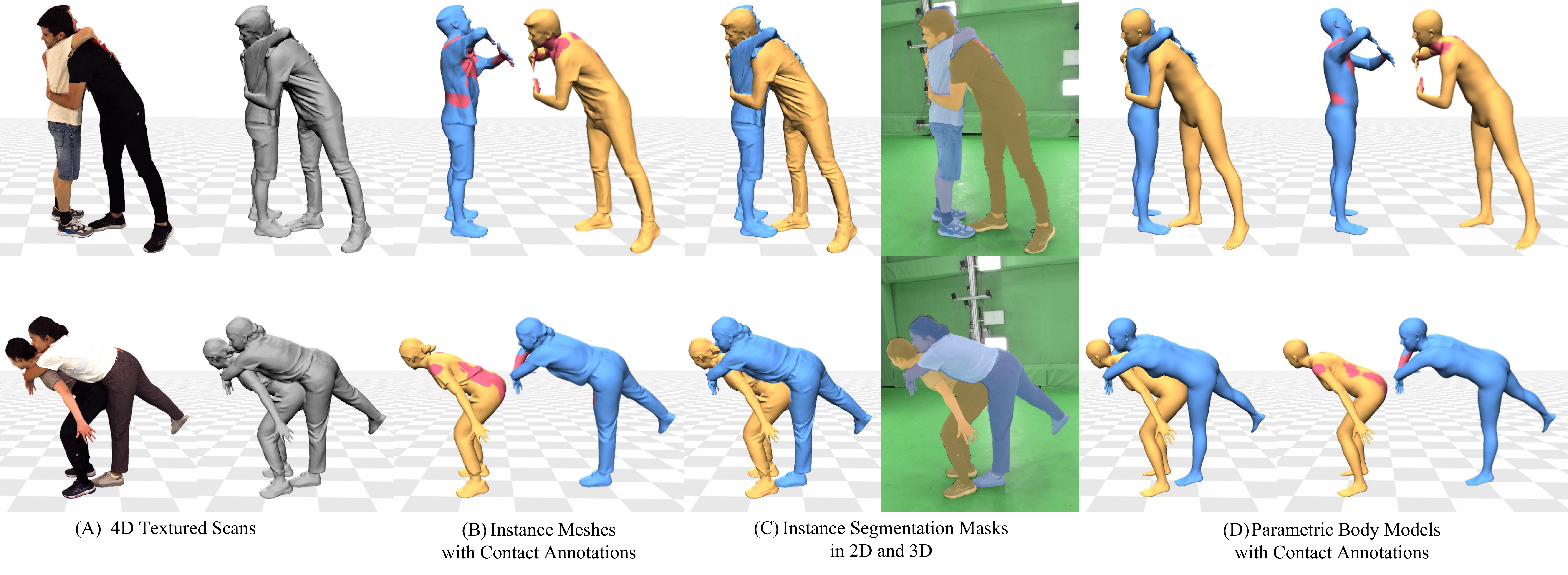}
\captionof{figure}{\textbf{Method and dataset.} We propose a method that leverages personalized human avatars to segment merged (A) 4D textured scans of multiple closely interacting humans. Based on the (B) instance meshes (with vertex-level contact annotations) we obtained in 3D space, we further provide (C) instance segmentation masks in 2D and 3D, (D) registered parametric body models with contact annotations.}
\label{fig: teaser}
\end{strip}


\begin{abstract}
We propose \datasetname, a method and dataset for the automatic analysis of physically close human-human interaction under prolonged contact. 
Robustly disentangling several in-contact subjects is a challenging task due to occlusions and complex shapes. Hence, existing multi-view systems typically fuse 3D surfaces of close subjects into a single, connected mesh.
To address this issue we leverage i) individually fitted neural implicit avatars; ii) an alternating optimization scheme that refines pose and surface through periods of close proximity; and iii) thus segment the fused raw scans into individual instances.
From these instances we compile \datasetname dataset of 4D textured scans of 20 subject pairs, 100 sequences, and a total of more than 11K frames. 
\datasetname contains rich interaction-centric annotations in 2D and 3D alongside accurately registered parametric body models.
We define varied human pose and shape estimation tasks on this dataset and provide results from state-of-the-art methods on these benchmarks. Project page: \url{https://yifeiyin04.github.io/Hi4D/}.

\end{abstract}


\section{Introduction}
\label{sec:intro}

\definecolor{LightGray}{gray}{0.9}
\begin{table*}[t]
\small
\resizebox{1.00\linewidth}{!}{
    \begin{tabular}{lcccccc}
    \hline
    \multirow{2}{*}{Dataset} & \multirow{2}{*}{Multi-view Images} & \multirow{2}{*}{Temporal} & \multicolumn{4}{c}{Reference Data Modalities} \\
    \cline{4-7} & & & 3D Pose Format & Textured Scans & Contact Annotations & Instance Masks \\
    \hline
    \rowcolor{LightGray}
    ShakeFive2 \cite{gemeren2016shakefive2} & & \checkmark & Joint Positions & & &\\
    MuPoTS-3D \cite{mehta2018mopots3d} & \checkmark & \checkmark & Joint Positions & & &\\
    \rowcolor{LightGray}
    ExPI \cite{guo2021expi}      & \checkmark &  \checkmark& Joint Positions & \checkmark & &\\
    MultiHuman \cite{zheng2021deepmulticap}  & & & Parametric Body Model$^\dag$ & \checkmark & &\\
    \rowcolor{LightGray}
    CHI3D \cite{fieraru2020chi} & \checkmark & \checkmark & Parametric Body Model & & region-level (631 events) &\\
    \hline
    \datasetname (Ours) & \checkmark &  \checkmark  & Parametric Body Model & \checkmark & vertex-level ($>$ 6K events) & \checkmark\\ \hline
    \end{tabular}
}
\caption{\textbf{Comparison of datasets containing close human interaction.} $^\dag$In \cite{zheng2021deepmulticap} registrations are not considered as ground-truth.}
\label{tab: dataset}
\end{table*}


%
While computer vision systems have made rapid progress in estimating the 3D body pose and shape of individuals
and well-spaced groups,
currently there are no methods that can robustly disentangle and reconstruct \emph{closely} interacting people. 
This is in part due to the lack of suitable datasets.
While some 3D datasets exist that contain human-human interactions, 
like ExPI \cite{guo2021expi} and CHI3D \cite{fieraru2020chi},
they typically lack high-fidelity dynamic textured geometry, do not always provide registered parametric body models and do not always provide rich contact information and are therefore not well suited to study closely interacting people. 

Taking a first step towards future AI systems that are able to interpret the interactions of multiple humans in close physical interaction and under strong occlusion, we propose a method and dataset that enables the study of this new setting.
Specifically, we propose \datasetname, a comprehensive dataset that contains segmented, yet complete 4D textured geometry of \textit{closely} interacting humans, alongside corresponding registered parametric human models, instance segmentation masks in 2D and 3D, and vertex-level contact annotations (see \cref{fig: teaser}). 
To enable research towards automated analysis of close human interactions, we contribute experimental protocols for computer vision tasks that are enabled by \datasetname.

Capturing such a dataset and the corresponding annotations is a very challenging endeavor in itself. 
While multi-view, volumetric capture setups can reconstruct high-quality 4D textured geometry of individual subjects, even modern multi-view systems typically fuse 3D surfaces of spatially proximal subjects into a single, connected mesh (see \figref{fig: teaser}, A).
Thus deriving and maintaining complete, per subject 4D surface geometry, parametric body registration, and contact information from such reconstructions is non-trivial.
In contrast to the case of rigid objects, simple tracking schemes fail due to very complex articulations and thus strong changes in terms of geometry. Moreover, contact itself will further deform the shape.

\begin{figure*}[t]
  \centering
    \includegraphics[width=\textwidth]{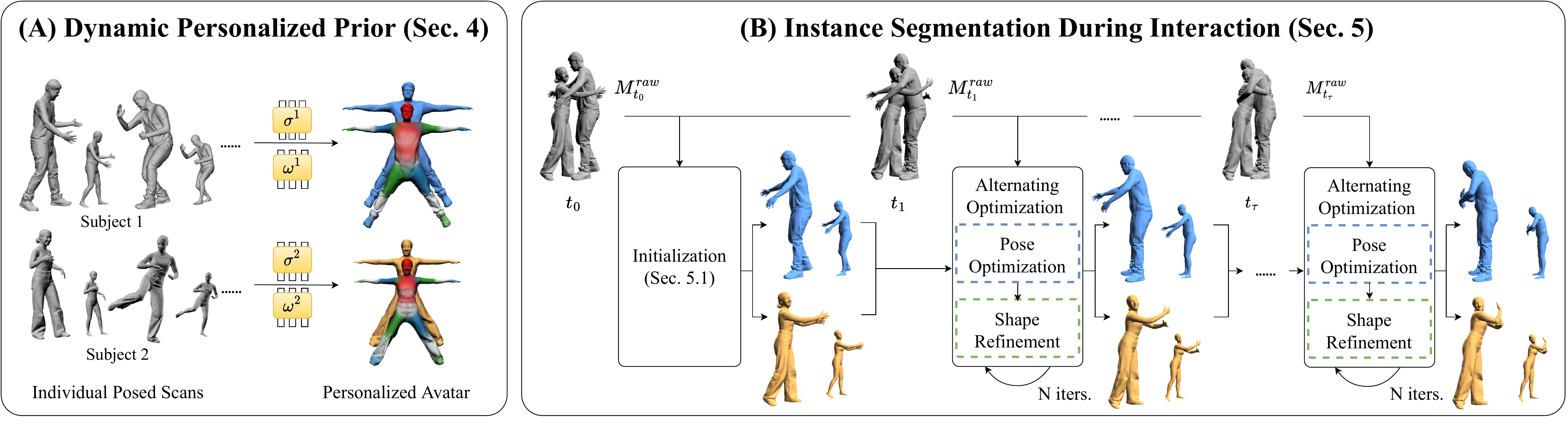}
  \caption{\textbf{Method overview.} (A) Dynamic Personalized Priors: We build individual personalized implicit avatars from 4D posed scans of each subject by modeling shape and deformation fields in canonical space following \cite{chen2021snarf}. (B) Instance Segmentation During Interaction: We then leverage the pre-built individual avatars to track and segment the raw 4D scans of multiple closely interacting people through extended periods of dynamic physical contact by optimizing pose and shape in an alternating manner (\cf \figref{fig: alternating_opt} for more details).
  }
  \label{fig: method}
\end{figure*}

To address these problems, we propose a novel method to track and segment the 4D surface of multiple closely interacting people through extended periods of dynamic physical contact. 
Our key idea is to make use of emerging neural implicit surface representations for articulated shapes, specifically SNARF \cite{chen2021snarf}, and create personalized human avatars of each individual (see \figref{fig: method}, A). These avatars then serve as strong personalized priors to track and thus segment the fused geometry of multiple interacting people (see \figref{fig: method}, B). To this end, we alternate between pose optimization and shape refinement  (see \figref{fig: alternating_opt}). The optimized pose and refined surfaces yield precise segmentations of the merged input geometry.
The tracked 3D instances (\figref{fig: teaser}, B) then provide 2D and 3D instance masks (\figref{fig: teaser}, C), vertex-level contact annotations (\figref{fig: teaser}, B), and can be used to register parametric human models (\figref{fig: teaser}, D). 

Equipped with this method, we capture \datasetname, which stands for \underline{H}umans \underline{i}nteracting in \underline{4D}, a dataset of humans in close physical interaction alongside high-quality 4D annotations. The dataset contains 20 pairs of subjects (24 male, 16 female), and 100 sequences with more than 11K frames. To our best knowledge, ours is the first dataset containing rich interaction-centric annotations and high-quality 4D textured geometry of closely interacting humans.

To provide baselines for future work, we evaluate several state-of-the-art methods for multi-person pose and shape modeling from images on \datasetname in different settings such as monocular and multi-view human pose estimation and detailed geometry reconstruction. Our baseline experiments show that our dataset provides diverse and challenging benchmarks, opening up new directions for research. 
\noindent In summary, we contribute:
\begin{compactitem}
    \item A novel method based on implicit avatars to track and segment 4D scans of closely interacting humans. 
    \item \datasetname, a dataset of 4D textured scans with corresponding multi-view RGB images, parametric body models, instance segmentation masks and vertex-level contact.
    \item Several experimental protocols for computer vision tasks in the close human interaction setting.
\end{compactitem}

\section{Related Work}
\label{sec:rw}

\noindent \textbf{Instance Segmentation.}
Most works that tackle human instance segmentation \cite{Lin2022RobustHV,Sun2022HumanIM,liu2011markerless, liu2013markerless,Seong2022OneTrimapVM} or object detection in general \cite{Girshick2015FastRCNN, Ren2015FasterRCNN, redmon2016yolo, he2017maskrcnn, bolya2019yolact, kirillov2020pointrend,Lei2022Transfiner,Liu2022GlobalSF} are only applicable to the 2D domain.
These methods do not na\"ively transfer to the 3D domain \cite{hou20193dsis}.
For 3D instance segmentation, previous work predominantly focuses on scene understanding that does not include humans \cite{dai2017scannet,Engelmann20CVPR}.
Thus, 3D instance segmentation of humans in close interactions is a relatively under-explored task.
Unlike static objects, humans undergo articulated motion, interact dynamically with surroundings, and cannot be represented by simple geometric primitives, which makes human-centric 3D instance segmentation inherently challenging.
To address this challenging problem, we create  personalized human avatars of each individual and subsequently leverage them as priors to track and segment closely interacting 3D humans.

\noindent\textbf{Human Pose and Shape Modeling.}
Explicit body models \cite{loper2015smpl,pavlakos2019expressive,Joo2018TotalCapture,xu2020ghum,Anguelov2005SCAPE} are widely used for human modeling in computer vision and computer graphics. Because of their low-dimensional parameter space and fixed topology of the underlying 3D mesh, they are well suited for learning tasks like fitting to RGB images \cite{kocabas2019vibe,kanazawa2018hmr,kolotouros2019spin,Li2021hybrik,sun2021romp,Choutas2020ExPose,pavlakos2019expressive}, RGB-D \cite{Bogo2015DetailedFR,Chen2016RealtimeRO,DoubleFusion}, or sparse point clouds \cite{Loper2014MoSh,AMASS:2019}.
Yet, the fixed 3D topology limits the maximum resolution and the expressive power to represent individual features and clothing.
While there have been efforts to alleviate this \cite{Alldieck2018SMPLclothing,alldieck2019learning, guo2021human, bhatnagar2019mgn,ma2020cape}, recent attention has turned to the use of implicit representations to tackle these limitations.
\cite{saito2021scanimate, deng2020nasa, tiwari2021neuralgif, chen2021snarf} have shown promising results for modeling articulated clothed human bodies in 3D. 
Among these, SNARF \cite{chen2021snarf} achieves state-of-the-art results and shows good generalization to unseen poses. We thus use it as a building block in our method, which - according to our ablations - is the more suitable choice than an explicit body model.

\noindent\textbf{Multi-Person Pose and Shape Estimation.}
Compared to the remarkable progress that has been made in estimating pose and shape of single humans from images or videos \cite{kanazawa2018hmr, joo2020eft, kolotouros2019spin, saito2019pifu, huang2020arch, zheng2021pamir, saito2020pifuhd, xiu2022icon, jiang2022selfrecon,Li2021hybrik, song2020human, guo2023vid2avatar}, not much attention has been paid to multi-person pose and shape estimation for \textit{closely} interacting humans. Multi-person estimators, \eg \cite{kocabas2019vibe, kocabas2021pare, jiang2020coherent, sun2021romp, sun2022bev, mustafa2021multi, dong2019mvpose, dong2021shapeaware, wang2021mvp}, mainly deal with the case where people are far away from each other and do not interact naturally in close range.
While the works of \cite{zhang2021lightweight,zheng2021deepmulticap} show more closely interacting people, the focus lies more on occlusions caused by this scenario and the actual contact between body parts is often limited.
\cite{fieraru2020chi} study closer human interactions in a similar setting to ours, but without textured scans. Their method to disentangle interacting people is fundamentally different from ours and heavily relies on manual annotations, which our method is able to avoid through the designed optimization schema.

\noindent\textbf{Close Human Interaction Datasets.}  There are several contact-related datasets focusing on how humans interact with objects or static scenes \cite{GRAB:2020, PROX:2019, bhatnagar22behave, RICH:2022, fan2023arctic}. None of them considers close interactions between dynamic humans.
Of the datasets containing human-human interactions \cite{mehta2018mopots3d, joo2015panoptic, hu2013k3hi, gemeren2016shakefive2, fieraru2020chi, zheng2021deepmulticap, guo2021expi} the most recent ones with close human interactions are summarized in \tabref{tab: dataset}. 
ShakeFive2\cite{gemeren2016shakefive2} and MuPoTS-3D \cite{mehta2018mopots3d} only provide 3D joint locations as reference data, lacking body shape information. The most related dataset to ours is CHI3D\cite{fieraru2020chi}, which employs a  motion capture system to fit parametric human models of at most one actor at a time. CHI3D only provides contact labels at body region-level and only for 631 frames, whereas we provide vertex-level annotations at more than 6K instances.  
Furthermore, CHI3D does not contain textured scans, which are crucial to evaluate surface reconstruction tasks. MultiHuman \cite{zheng2021deepmulticap} provides textured scans of interacting people, but only of 453 static frames and without ground-truth level body model registrations. ExPI \cite{guo2021expi} contains dynamic textured meshes in addition to 3D joint locations, but misses instance masks along with body model registrations and contact information. Moreover, there are only two pairs of dance actors in ExPI \cite{guo2021expi}, thus lacking in subject and clothing diversity.
In contrast, our dataset \datasetname encompasses a rich set of data modalities for closely interacting humans.


\section{Approach Overview}
\vspace{-0.2em}
Vision-based disentanglement of in-contact subjects is a challenging task due to strong occlusions and a priori unknown geometries. 
Hence, multi-view systems typically fuse 3D surfaces of close subjects into a single, connected mesh.
Here we detail our method to segment 4D scans of closely interacting people to obtain instance-level annotations. 
Our method makes use of three components: 
i) we fit individual neural implicit avatars to frames without contact (\cf \figref{fig: method}, A \& \cref{subsec: prior_intro}); 
ii) these serve as personalized priors in an alternating optimization scheme that refines pose (\cref{subsec: pose_opt}) and surface (\cref{subsec: shape_refine}) through periods of close proximity; and 
iii) thus segment the fused 4D raw scans into individual instances (\cf \figref{fig: method}, B \& \cref{sec: method}).


\section{Dynamic Personalized Prior}
\vspace{-0.2em}
\label{subsec: prior_intro}

\begin{figure*}[!ht]
  \centering
    \includegraphics[width=\textwidth]{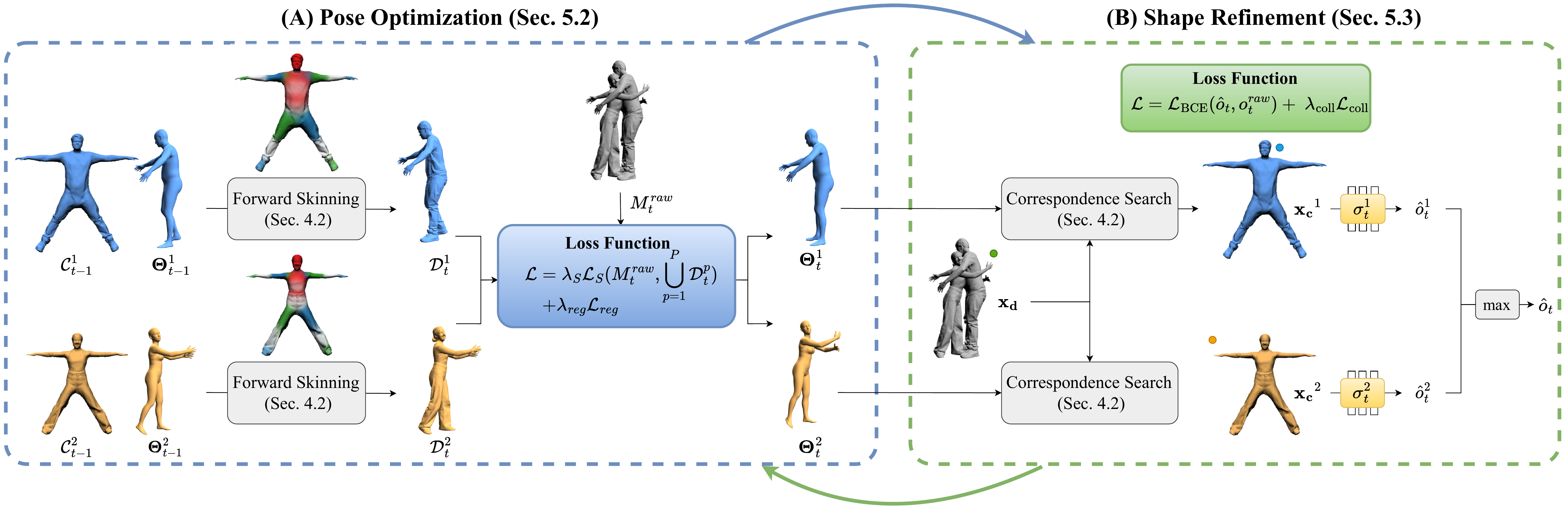}
  \caption{\textbf{Alternating optimization.} (A) Given personalized avatars for each subject (\figref{fig: method}), we jointly optimize the poses of each subject $\boldsymbol{\Theta}^p_t$ using a surface energy term. (B) With $\boldsymbol{\Theta}^p_t$ and merged raw scans $M^{raw}_t$, we refine the shape network weights $\sigma^p_t$ of the individual avatars on the fly to maximally preserve the details and model contact-aware deformations. We alternate between optimizing (A) and (B) for $N$ iterations. Afterwards, the optimized $\boldsymbol{\Theta}^p_t$ and refined $\sigma^p_t$ serve as initialization for the optimization process at the next frame $t+1$.}
  \label{fig: alternating_opt}
\end{figure*}

To build personalized priors, we first capture 4D scans for each subject during dynamic motion. Minimally clothed parametric body models (here, SMPL) are registered to these scans (\secref{subsec: fit_smpl}). Next, detailed avatars are learned for each subject (\secref{subsec: snarf}). We leverage these to alleviate instance ambiguities during close interaction  (\secref{sec: method}).

\subsection{Parametric Body Model Fitting}
\label{subsec: fit_smpl}
We register SMPL \cite{loper2015smpl} to the individual scans to represent the underlying body and its pose. 
SMPL is defined as a differentiable function 
that maps shape parameters $\beta \in \mathbb{R}^{10}$, pose parameters $\theta \in \mathbb{R}^{72}$ and translation $t \in \mathbb{R}^3$ to a body mesh $\mathcal{M}$ with $6890$ vertices.
Registering the SMPL model is formulated as an energy minimization problem over body shape, pose, and translation parameters:
\begin{equation}
    E(\beta, \theta, t)=\lambda_S E_S+\lambda_J E_J+\lambda_\theta E_\theta+\lambda_\beta E_\beta \ ,
    \label{eq: fit_smpl}
\end{equation}
where $E_S$ denotes the bi-directional distances between the SMPL mesh and the corresponding scan, and $E_J$ is a 3D keypoint energy between regressed SMPL 3D joints and 3D keypoints which are obtained via triangulation of 2D keypoints detected in the multi-view RGB images~\cite{cao2017openpose}. 
$E_\theta$ and $E_\beta$ are prior terms to constrain human pose and shape (\cf~\cite{bogo2016simplify}).
Each term is weighted with a corresponding weight $\lambda$. See Sup. Mat. for more details.

To obtain high-quality registrations, the body shape $\beta$ is estimated in advance from a minimally clothed static scan for each subject following \cite{bhatnagar2020ipnet, bhatnagar2020loopreg}. For registering SMPL to clothed scans, we keep $\beta$ fixed in \equref{eq: fit_smpl}. For brevity, we summarize the parameters as $\boldsymbol{\Theta} =  (\beta, \theta, t)$.

\subsection{Human Avatar Learning}
\label{subsec: snarf}
Neural implicit surfaces can be used to represent articulated human bodies \cite{saito2021scanimate, deng2020nasa, tiwari2021neuralgif, chen2021snarf}. We leverage SNARF~\cite{chen2021snarf} for its generalization to unseen and challenging poses.
Following \cite{chen2021snarf} we use two neural fields to model shape and deformation in canonical space: 
\begin{itemize}
  \item \textbf{Shape Field:}  $f_{\sigma}$ is used to predict the occupancy probability $\hat{o}\left(\mathbf{x}_c, \boldsymbol{\Theta}\right)$ of any 3D point $\boldsymbol{x}_c$ in canonical space, where 1 is defined as inside and 0 as outside. The SMPL pose parameters are provided as an input to model pose-dependent deformations. The canonical shape is implicitly defined as the 0.5 level set $\mathcal{C} = \{ \boldsymbol{\ x}_c \ |\ f_{\sigma}(\boldsymbol{x}_c,\boldsymbol{\Theta}) = 0.5 \ \}$.
  \item  \textbf{Deformation Field:} $\mathbf{w}_{\omega}$ denotes a person-specific, canonical deformation field. It transforms the acquired shape to a desired pose $\boldsymbol{\Theta}$ via linear blend skinning (LBS) with learned deformation field.
\end{itemize}
\textbf{Correspondence Search.} \label{subsubsec: correspondence} Given a query point sampled in deformed space $\boldsymbol{x}_d$, \cite{chen2021snarf} determines its correspondence $\boldsymbol{x}_c$ in canonical space via iterative root finding such that it satisfies the forward skinning function $\boldsymbol{x}_d=\boldsymbol{d}_{\omega}(\boldsymbol{x_c}, \boldsymbol{\Theta})$.  
 
\noindent\textbf{Training Losses.} The implicit model is then trained via binary cross entropy $\mathcal{L}_{B C E}\left(\hat{o}\left(\mathbf{x}_d,\boldsymbol{\Theta}\right), o^{raw}\left(\mathbf{x}_d \right)\right)$, formed between the predicted occupancy $\hat{o}\left(\mathbf{x}_d, \boldsymbol{\Theta}\right)$ and the ground-truth occupancy $o^{raw}\left(\mathbf{x}_d \right)$ of points $\mathbf{x}_d$ in deformed space.

\noindent\textbf{Mesh Extraction.} We use \textit{Multiresolution IsoSurface Extraction} (MISE) \cite{mescheder2019occupancy} to extract meshes $\mathcal{C}$ from the continuous occupancy fields in canonical space: 
\begin{equation}
    \mathcal{C} = MISE(f_{\sigma}, \boldsymbol{\Theta}).
    \label{eq: MISE}
\end{equation}
The canonical shape can be deformed to posed space via linear blend skinning using the learned deformation field $\mathbf{w}_{\omega}$. For brevity, we denote this deformation as 
\begin{equation}
    \mathcal{D} = LBS(\mathcal{C}, \mathbf{w}_{\omega}, \boldsymbol{\Theta}).
    \label{eq: LBS}
\end{equation}


\section{Instance Segmentation During Interaction}
\label{sec: method}

\begin{algorithm}[t]
\small
\caption{Alternating optimization to estimate SMPL parameters $\boldsymbol{\Theta}^{p}_t$ and refine shape network weights $\sigma^{p}_t$ for each frame $t_1 \leqslant t \leqslant t_\tau$ in which contact happens.}
\label{alg: algorithm}
\begin{algorithmic}
\State $\boldsymbol{\Theta}^{p}_{t_0} \gets$ \text{Initial SMPL parameters of subject $p$ at $t_0$} 
\State ${\sigma}^{p}_{t_0} \gets$ \text{Initial shape network weights of subject $p$ at $t_0$} 
\For{$t = t_1, \dots, t_{\tau}$}
\State${\boldsymbol{\Theta}^{p}_t}^{(0)} \gets {\boldsymbol{\Theta}^{p}_{t-1}}$
\State${{{\sigma}^{p}_t}^{(0)} \gets {\sigma}^{p}_{t-1}}$
\For{$n = 1, \dots, N$}
\State ${\boldsymbol{\Theta}^{p}_t}^{(n)} \gets$ \text{Pose Optimization (\equref{eq: pose_opt}, \secref{subsec: pose_opt})}
\State ${{\sigma}^{p}_t}^{(n)} \gets$ \text{Shape Refinement (\equref{eq: shape_refine}, \secref{subsec: shape_refine})}
\EndFor
\State ${\boldsymbol{\Theta}^{p}_t} \gets {\boldsymbol{\Theta}^{p}_t}^{(N)}$
\State ${{\sigma}^{p}_t} \gets {{\sigma}^{p}_t}^{(N)}$
\EndFor
\end{algorithmic}
\end{algorithm}
Given pre-built individual implicit avatars obtained in  \secref{subsec: prior_intro}, our goal is to track and segment the 4D scans through extended periods of dynamic physical contact. 
To this end, we leverage the avatars as priors to compensate for ambiguities caused by contact.
The process includes the following steps: 
i) given the last frame $t_0$ before contact, we initialize pose parameters from the still separated scans (\secref{subsec: init}); 
ii) starting from the first frame with contact, we then jointly refine the pose parameters $\boldsymbol{\Theta}_t^p$ for all $P$ subjects $p \in \{1,\dots,P\}$ in the scene via minimization of a surface energy term (\secref{subsec: pose_opt}); iii) we further refine the implicit shape network weights ${\sigma}_t^p$ on the fly throughout the interaction sequence, using the optimized poses and raw scans which at this point are merged. This leads to maximal preservation of details and allows modelling of contact-aware deformations (\secref{subsec: shape_refine}). Steps ii) and iii) are performed in an alternating fashion for $N$ steps. To be noted, this is a tracking process over time. The method is illustrated in \figref{fig: alternating_opt} and Alg.~\ref{alg: algorithm}. 

\subsection{Initialization}
\label{subsec: init}
We denote the last frame without physical contact by $t_0$ and denote the last frame with contact by $t_\tau$. 
We register the SMPL model to separated scans to obtain the initial pose parameters $\boldsymbol{\Theta}_{0}^{p}$ for each subject. 
We further use $\boldsymbol{\Theta}_{0}^{p}$ and the corresponding avatar to extract the canonical shape $\mathcal{C}_{0}^p$ for each subject $p$ via \equref{eq: MISE}.
During frames $t \in \{t_1, \dotsc, t_\tau \}$ with physical contact, the raw scan ${M_{t}^{raw}}$ is fused together. To track through this period, we initialize the shape $\mathcal{C}_t^p$ for each subject from the last frame $\boldsymbol{\Theta}_{t-1}^{p}$.

\subsection{Pose Optimization}
\label{subsec: pose_opt}
To obtain the SMPL parameters $\boldsymbol{\Theta}_{t}^{p}$ for contact frames, we optimize the following objective:
\begin{equation}
        \mathcal{L} =\lambda_{s2m} \mathcal{L}_{s2m}({M_{t}^{raw}},\bigcup_{p=1}^{P}  \mathcal{D}_t^p) + \lambda_{\text {reg}}\mathcal{L}_{\text {reg}} , 
\label{eq: pose_opt}
\end{equation}
where $\mathcal{L}_{s2m}$ encourages the union of the posed meshes $\mathcal{D}_t^p = LBS(\mathcal{C}_{t}^p, \mathbf{w}_{\sigma_w}, \boldsymbol{\Theta}_{t}^{p})$ (\equref{eq: LBS}) to align with the fused input scan ${M_{t}^{raw}}$ by optimizing the SMPL parameters of each subject $\boldsymbol{\Theta}_{t}^{p}$ jointly.
$\mathcal{L}_{\text {reg}}$ is a regularization term:
\begin{equation}
    \mathcal{L}_{\text{reg}} = \sum^{P}_{p=1} \mathcal{L}_{s2m}(\mathcal{D}_t^p, {\mathcal{M}}_{t}^{p}) + \lambda_{\Theta}\mathcal{L}_{\Theta} (\boldsymbol{\Theta}_{t}^{p}) .
    \label{eq: pose_reg}
\end{equation}
The term $\mathcal{L}_{s2m}(\mathcal{D}_t^p, {\mathcal{M}}^{p})$ 
ensures that the SMPL template $\mathcal{M}^{p}$ aligns well with each subject's deformed surface and $\mathcal{L}_{\Theta}$ is a prior penalizing unrealistic human poses (\cf~\cite{bogo2016simplify}) and $\lambda_{(\cdot)}$ denote the corresponding weights. The scan-to-mesh loss term $\mathcal{L}_{s2m}$ is defined in Supp. Mat.

\subsection{Shape Refinement}
\label{subsec: shape_refine}
After the pose optimization stage, we refine the shape networks $f_{\sigma_{t}^{p}}$ of each avatar  to retain high-frequency details and to model contact-induced deformations. 
To achieve this, we sample points $\boldsymbol{x}_d$ on the raw scan ${M_{t}^{raw}}$ and find canonical correspondences $\boldsymbol{x}^{p}_c$ per subject, given the optimized poses $\boldsymbol{\Theta}_{t}^{p}$. For each $\boldsymbol{x}_c$, the shape network $f_{\sigma_{t}^{p}}$ predicts the subject-specific occupancy $\hat{o}_{t}^{p}$ (\cref{subsubsec: correspondence}). The final occupancy prediction $\hat{o}_{t}$ of the sampled query point $\boldsymbol{x}_d$ is composited as the union over the individual predictions $\hat{o}_{t}^{p}$:
\begin{equation}
    \hat{o}_{t} = \max_{p\in \{1,\dots, P\}}[\hat{o}_{t}^{p}] =\max_{p\in \{1,\dots, P\}}[f_{\sigma_{t}^{p}}(\boldsymbol{x}_d,\boldsymbol{\Theta}_{t}^p)].
\end{equation}
We then refine the shape network weights $\sigma_{t}^{p}$ of the avatars by minimizing the loss: 
\begin{equation}
\mathcal{L} = \mathcal{L}_{\text{BCE}}(\hat{o}_{t}, {{o}}_{t}^{raw})+ \lambda_{\text{coll}} \mathcal{L}_{\text{coll}}
\label{eq: shape_refine}
\end{equation}
where $\mathcal{L}_{\text{BCE}}(\hat{o}_{t}, {{o}}_{t}^{raw})$ is the binary cross entropy between the composited occupancy prediction and the corresponding point ${{o}}_{t}^{raw}$ on the input  scan. This encourages segmented avatars to, together, align well with the fused scan.

A key challenge is to correctly model contact-induced deformation. Recall that the avatars are initialized from non-contact frames and hence do not yet account for the flattening of clothing and soft tissue due to contact. Therefore, the pose optimization step will cause surfaces that are in contact to intersect.     
To alleviate this, we select points that are predicted to be inside multiple subjects by querying the individual occupancies. 
We denote this subset of points as $\mathcal{S} = \{ \boldsymbol{x}_d \mid \hat{o}_{t}^{i}(\boldsymbol{x}_d) > 0.5, \hat{o}_{t}^{j}(\boldsymbol{x}_d) > 0.5 \ \forall \ i \in \{1,\dots,P\}, j \in \{1,\dots,P\}, i \neq j  \}$.
We then penalize interpenetration of surfaces via $\mathcal{L}_{\text{coll}}$: 
\begin{equation}
    \mathcal{L}_{\text{coll}}= \dfrac{1}{|{\mathcal{S}}|}\sum_{ \boldsymbol{x}_d \in \mathcal{S}} \phi(\hat{o}_{t}^{i}(\boldsymbol{x}_d)) \cdot \phi(\hat{o}_{t}^{j}(\boldsymbol{x}_d)) ,
    \label{eq: collision}
\end{equation}
where $\phi(\hat{o}_{t}(\boldsymbol{x}_d)) = \max(\hat{o}_{t}(\boldsymbol{x}_d)-0.5, 0)$. 
Intuitively, we ask that the uniformly sampled 3D points do not result in occupancy values of $1$ (i.e., inside) for multiple shapes simultaneously. Instead, the shape networks are optimized such that the surfaces adhere to contact deformation.

\section{Dataset} 
\label{sec:dataset}

We believe that with \datasetname we contribute a valuable tool for the community working on human-to-human close interaction.
For a detailed list of its contents and comparison to existing datasets, please refer to \tabref{tab: dataset} and Supp. Mat.

\begin{figure}[h]
  \centering
    \includegraphics[width=0.3\textwidth]{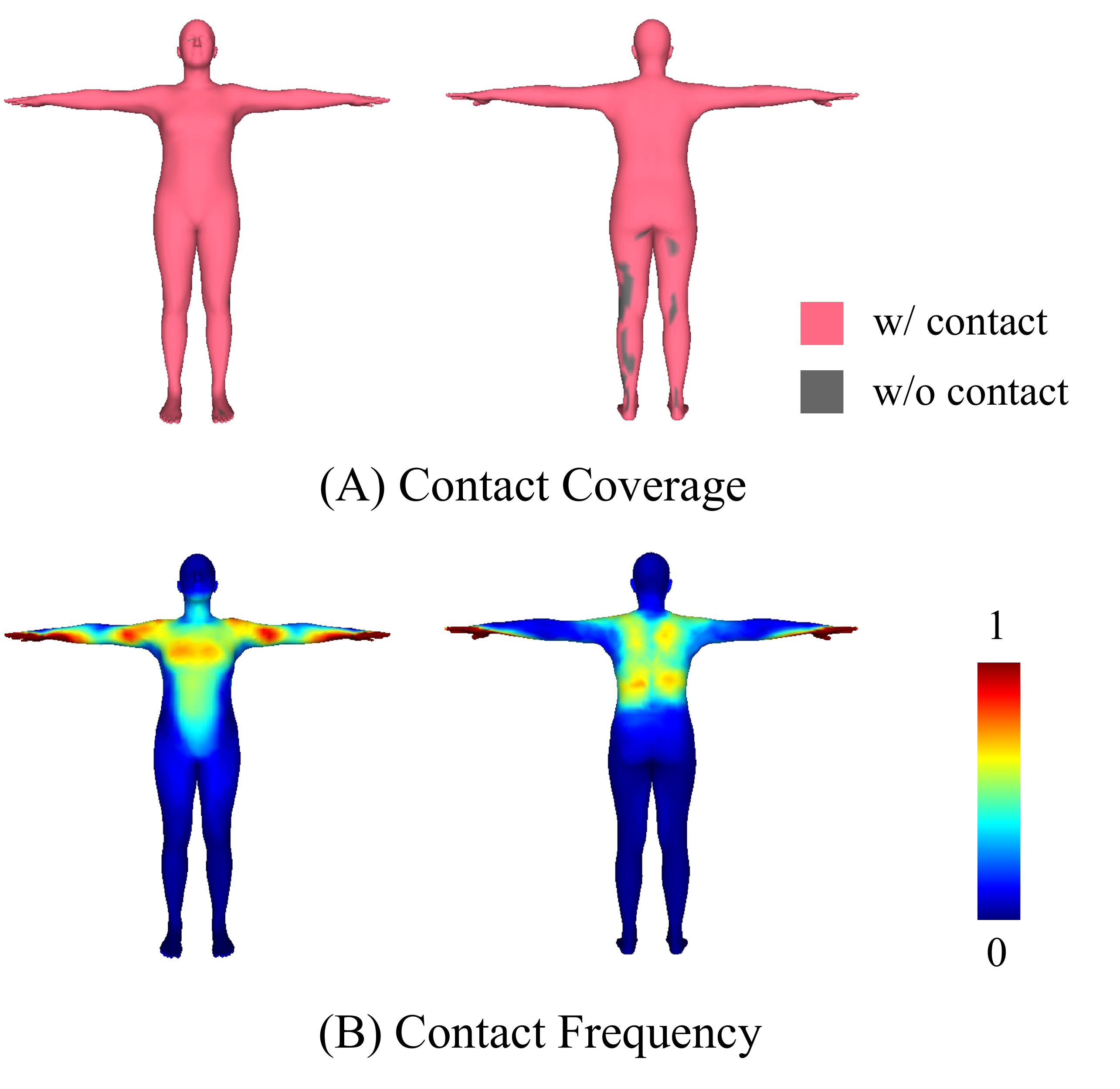}
  \caption{\textbf{Contact coverage and frequency of \datasetname.}}
  \label{fig: contact}

\end{figure}

We recruited 20 unique pairs of participants with varying body shapes and clothing styles to perform diverse interaction motion sequences of a few seconds, such as hugging, posing, dancing, and playing sports. We collected 100 independent clips with 11K frames in total and more than 6K frames of them with physical contact. Contact annotations in \datasetname cover over 95 \% of the parametric human body. The contact coverage and contact frequency of \datasetname is shown in \figref{fig: contact}.


\section{Experiment}
\label{sec:experiment}\label{subsec: mesh_eval}

We conduct ablations on \datasetname to verify our design choices (\secref{subsec:ablation}) and to compare with baselines (\secref{subsec:comparison}).
We consider the following metrics for reconstruction evaluation: volumetric IoU (mIoU) [\%], Chamfer distance ({C-}$L_2$) [cm], point-to-surface distance (P2S) [cm], and normal consistency (NC) [\%]. 

\begin{table}[!h]
  \centering
  \resizebox{1.00\linewidth}{!}{
  \begin{tabular}{lcccc}
    \toprule
     \textbf{Method} & \textbf{IoU}  $\uparrow$   &\textbf{C-}$\mathbf{L_2}$ $\downarrow$& \textbf{P2S} $\downarrow$ & \textbf{NC}  $\uparrow$   \\
         \midrule
     Ours (w/o shape refine)  & $0.982$   &  $0.36$   & $0.37$ & $0.934$   \\
     Ours (w/o alternating opt.)  & $0.938$   &  $0.48$   & $0.46$ & $0.927$   \\
     \midrule
     SMPL+D  & $0.983$          & $0.24$            & $0.30$          & $0.927$  \\
    \midrule
    Ours  & $\mathbf{0.989}$ & $\mathbf{0.22}$   & $\mathbf{0.23}$ & $\mathbf{0.945}$ \\
    \bottomrule
  \end{tabular}
  }
  \caption{\textbf{Quantitative Results.} Ablations to evaluate our method without the shape refinement stage and without alternating optimization and comparison to the SMPL+D baseline.}
  \label{tab: quant_exp}
\end{table}
\subsection{Ablation Study}
\label{subsec:ablation}

\begin{figure}[!h]
  \centering
    \includegraphics[width=0.4\textwidth]{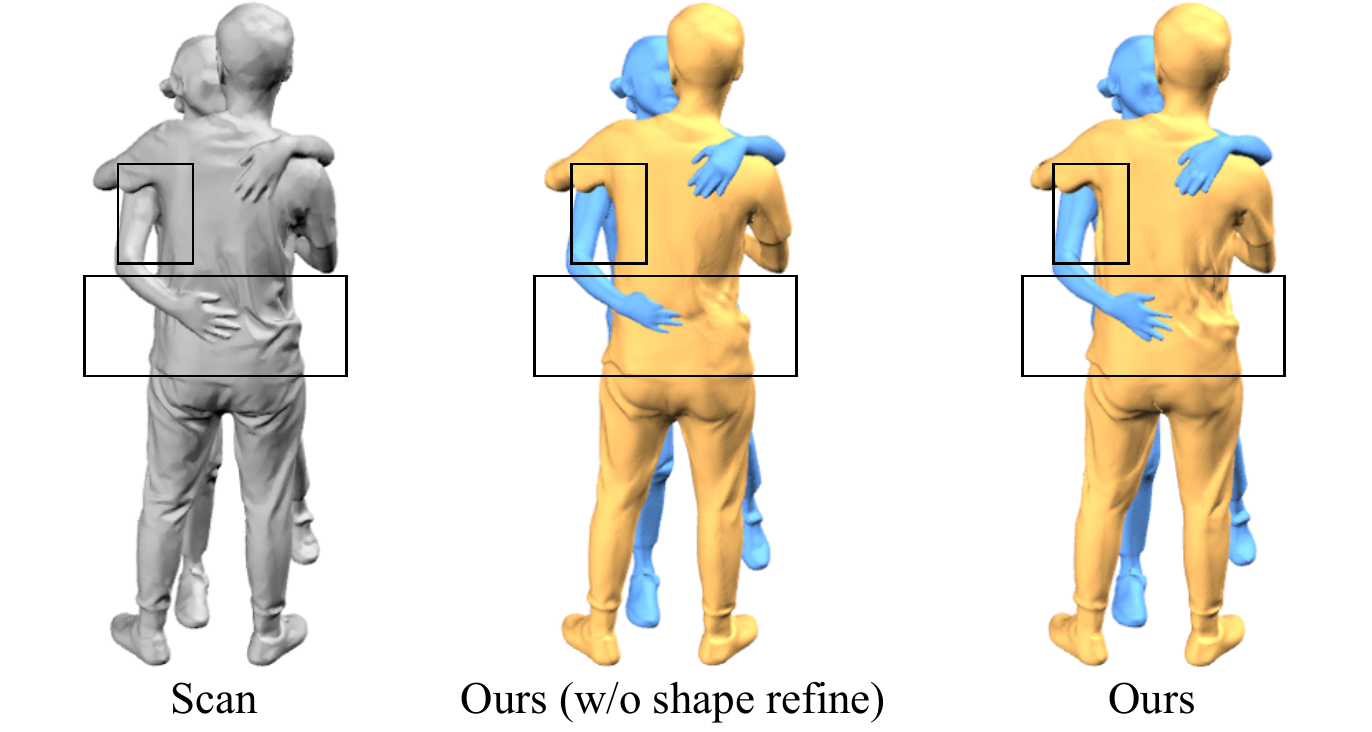}
  \caption{\textbf{Qualitative ablation (shape refinement).} The shape refinement stage better models the contact-aware deformation.}
  \label{fig: pose_only}
\end{figure}

\begin{figure}[!h]
  \centering
    \includegraphics[width=0.4\textwidth]{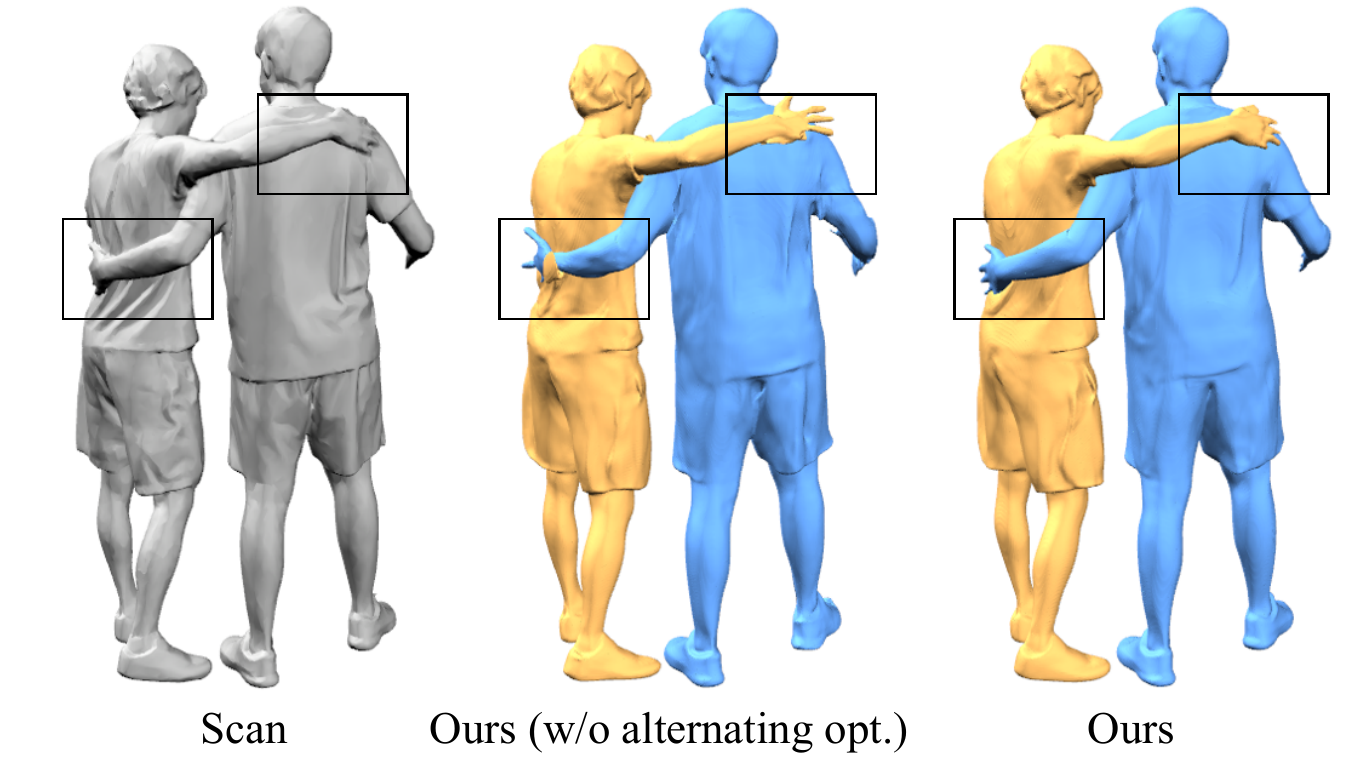}
  \caption{\textbf{Qualitative ablation (alternating optimization).} Optimizing poses and shape networks alternatingly improves results in areas with heavy contact.}
  \label{fig: joint_opt}
\end{figure}

\begin{figure}[!h]
  \centering
    \includegraphics[width=0.43\textwidth]{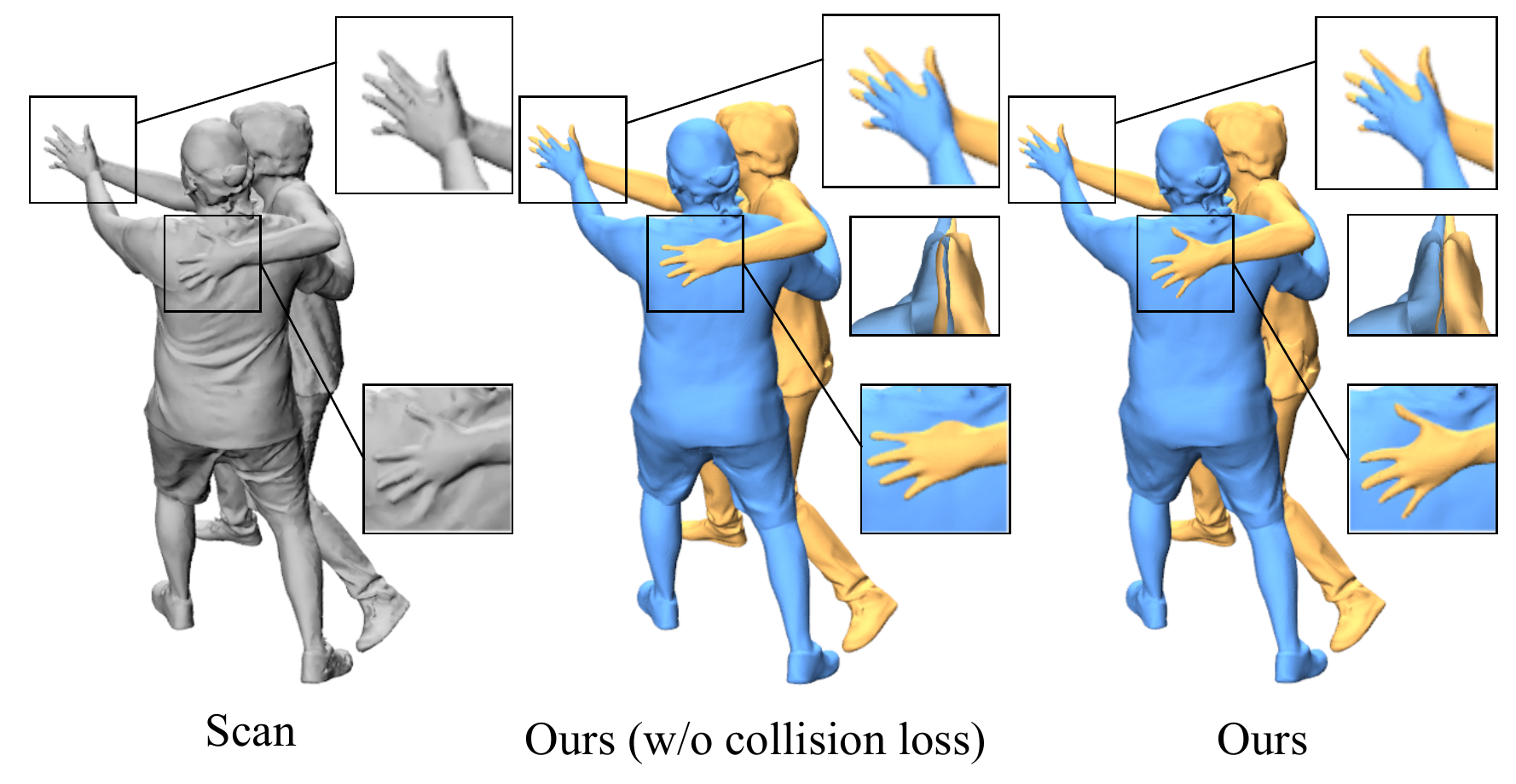}
  \caption{\textbf{Importance of collision loss.} 
  Instance meshes intersect each other in contact areas if we remove the collision loss term.}
  \label{fig: ablation_interp}
  \vspace{-2mm}
\end{figure}

\begin{table*}[t]
\centering
\small
\begin{tabular}{ccccccccc}
\hline
\textbf{Setting}  & \textbf{Method}  & \textbf{MPJPE}  $\downarrow$ & \textbf{MVE} $\downarrow$ & \textbf{NMJE} $\downarrow$ & \textbf{NMVE} $\downarrow$ & \textbf{F1} $\uparrow$& $\textbf{PCDR}^{0.10}$ $\uparrow$& \textbf{CD} $\downarrow$\\ \hline
\multicolumn{1}{c}{\multirow{3}{*}{Monocular}} & PARE   \cite{kocabas2021pare}          &   $87.6$    &  $106.5$   &  $95.3$    &    $115.9$  &  $0.919$  &   $0.610$   &      $297.7$         \\
\multicolumn{1}{c}{}                           & ROMP  \cite{sun2021romp}           &   $93.0$    &  $116.2$    &  $93.2$    &  $116.4$    &  $0.998$  &   $0.613$   &      $338.0$         \\
\multicolumn{1}{c}{}                           & BEV    \cite{sun2022bev}          &   $92.5$    &  $113.7$   &   $92.6$   &    $113.8$   &  $0.999$    &  $0.745$   &      $295.8$         \\ \hline
\multicolumn{1}{c}{\multirow{2}{*}{Multi-view}} & MVPose (4-views) \cite{dong2019mvpose} &   $61.3$     &   $78.3$  &    $67.0$   &   $85.4$    &  $0.917$  &   $0.957$   &    $234.8$           \\
\multicolumn{1}{c}{}    & MVPose (8-views) \cite{dong2019mvpose} &   $50.3$     &   $61.8$  &    $51.8$   &   $63.6$    &  $0.971$  &   $0.972$   &    $166.8$           \\
\hline
\end{tabular}
\caption{\textbf{SMPL estimation.} Results of monocular and multi-view SMPL estimation methods on \datasetname  (\cf Sec. \ref{subsec: smpl_baseline} and Fig. \ref{fig: smpl_estimation}).}
\label{tab: smpl_estimation}
\end{table*}

\begin{table}[h]
  \centering
  \resizebox{1.00\linewidth}{!}{
  \begin{tabular}{cccccc}
    \hline
      \textbf{Setting} & \textbf{Method} & \textbf{IoU}  $\uparrow$   &\textbf{C-}$\mathbf{L_2}$ $\downarrow$& \textbf{P2S} $\downarrow$ & \textbf{NC}  $\uparrow$    \\
    \hline
   \multicolumn{1}{c}{\multirow{2}{*}{Monocular}} & PIFuHD \cite{saito2020pifuhd}  & $0.761$ & $3.02$ & $2.89$ & $0.755$   \\
    \multicolumn{1}{c}{\multirow{2}{*}{}} & ICON \cite{xiu2022icon} & $0.780$   & $2.76$   & $2.54$ & $0.762$   \\
    \hline
    \multicolumn{1}{c}{\multirow{2}{*}{Multi-view}} & DMC \cite{zheng2021deepmulticap} (4-views)  & $0.893$    & $1.78$   & $1.82$ & $0.832$   \\
   \multicolumn{1}{c}{\multirow{2}{*}{}} &  DMC \cite{zheng2021deepmulticap} (8-views)   & $0.906$   & $1.64$   & $1.60$ & $0.851$   \\
     \hline
  \end{tabular}
  }
  \caption{\textbf{Detailed geometry reconstruction.} Results of monocular and multi-view detailed geometry reconstruction methods on \datasetname (\cf Sec. \ref{subsec: mesh_baseline} and Fig. \ref{fig: mesh_reconstruction}).}
  \label{tab:mesh_reconstruction}
  \vspace{-2mm}
\end{table}

\noindent\textbf{Shape Refinement.}
We compare our full optimization pipeline to a version without the shape refinement stage (\secref{subsec: shape_refine}). From \figref{fig: pose_only}, we observe that the details in the cloth are not accurately modelled if we do not further refine the shape network weights. Moreover, without the shape refinement stage, the method fails to model the contact-aware cloth deformations (\eg the right hand of the blue colored person is occluded by the other person's cloth). Note that such deformations only occur with physical contact between people and cannot be learned from individual scans.
The quantitative results in \tabref{tab: quant_exp} also support the benefits of the shape network refinement.

\noindent \textbf{Alternating Optimization.}
 We compare our alternating optimization with an approach where poses and shape networks are optimized concurrently.
We observe that in this case, it is hard to disambiguate body parts of different subjects in the contact area (\cf \figref{fig: joint_opt}).
Our alternating pipeline can better disentangle the effects of the pose and shape network.
The quantitative results of \tabref{tab: quant_exp} also confirm this.

\noindent\textbf{Collision Loss.}
Without penalizing the collision of the two occupancy fields, one person’s mesh might be partially intersected by the other person in the contact area, as we see from \figref{fig: ablation_interp}. 
We quantitatively measure the interpenetration by calculating the intersection volume between the individual segmented meshes. With the collision loss term defined in \equref{eq: collision}, the average intersection volume decreases from \SI{11.62e-4}{\meter^3} to \SI{5.82e-4}{\meter^3} by $49.91 \%$.

\subsection{Comparison Study}
\label{subsec:comparison}

\noindent\textbf{SMPL+D Baseline.} 
A straightforward baseline for our task is to directly track multiple clothed SMPL body template meshes. We define this baseline as the SMPL+D (\cf \cite{bhatnagar2020ipnet, bhatnagar2020loopreg}) tracking baseline. It tracks the 3D geometry of close interacting people at each frame by estimating the individual displacement of each SMPL vertex and each subject. 
We optimize these displacement fields and the SMPL body parameters in a similar alternating manner as we do in our proposed method.
For more implementation details please refer to the Supp. Mat.
Quantitatively, our proposed method with personalized priors outperforms the SMPL+D baseline on all metrics (\tabref{tab: quant_exp}). The optimized SMPL+D models do not have any personalized prior knowledge from which we can infer when substantial instance ambiguity exists. 
We show qualitative results in the Supp. Mat.


\section{Benchmark Baselines}

\begin{figure*}[]
  \centering
    \includegraphics[width=\textwidth]{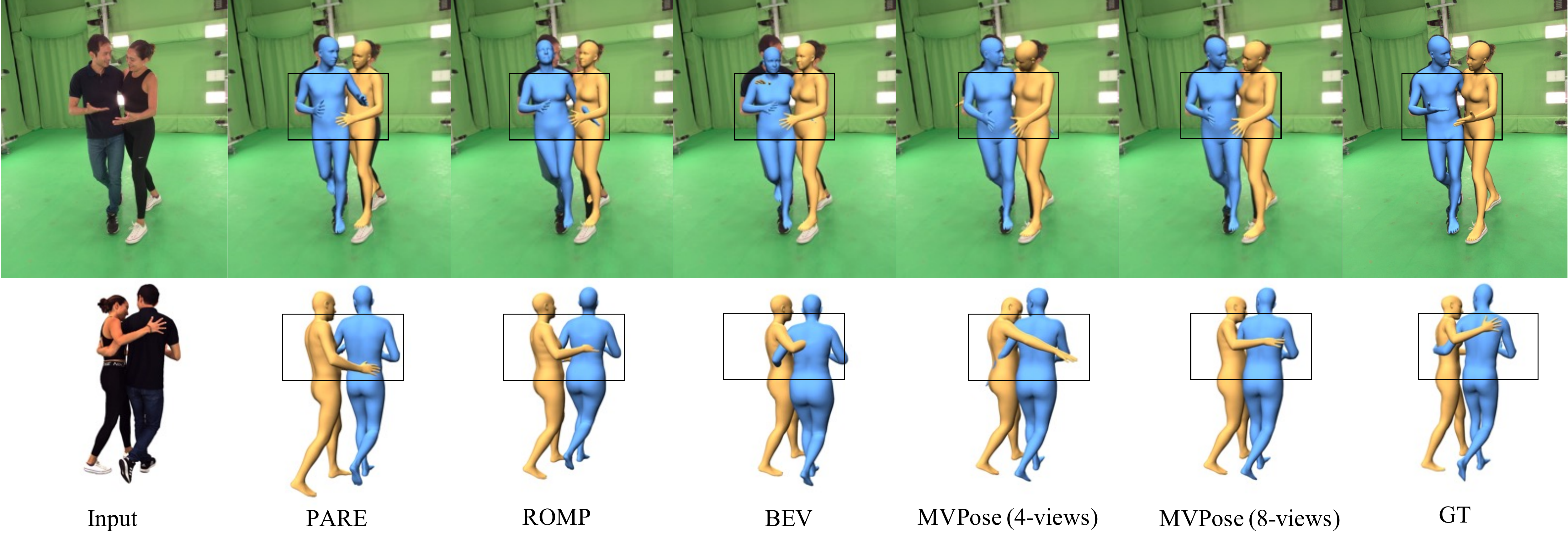}
  \caption{\textbf{SMPL estimation.} First row: results in 2D image space. Second row: results in an alternative 3D view (\cf Sec. \ref{subsec: smpl_baseline}).}
  \label{fig: smpl_estimation}
\end{figure*} 

\begin{figure*}[]
  \centering
    \includegraphics[width=\textwidth]{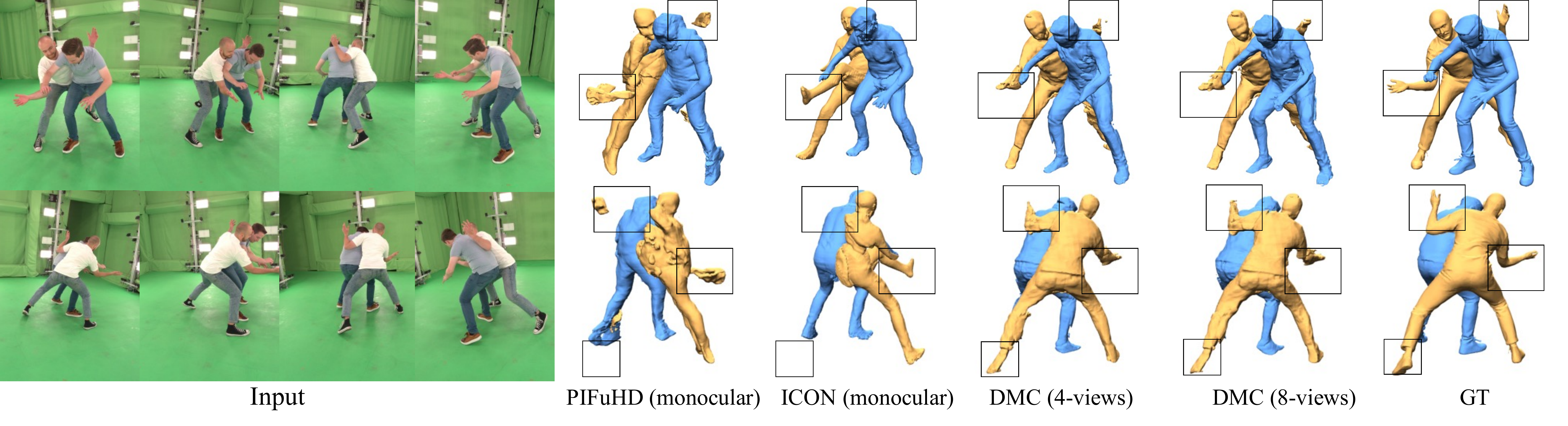}
  \caption{\textbf{Detailed geometry reconstruction.} 
  Results of monocular and multi-view methods together with the GT of \datasetname (\cf Sec. \ref{subsec: mesh_baseline}).
  }
  \label{fig: mesh_reconstruction}
\end{figure*} 

We define several standard vision benchmarks conducted on the \datasetname dataset. These benchmarks include monocular SMPL estimation, multi-view SMPL estimation, monocular detailed geometry reconstruction and multi-view detailed geometry reconstruction. 
We evaluate several baseline methods on each of these tasks and demonstrate experimentally that our \datasetname dataset is challenging, thus opening many doors for future research.

\subsection{SMPL Estimation}
\label{subsec: smpl_baseline}

\noindent\textbf{Evaluation Protocol.} 
\datasetname provides multi-view RGB sequences with corresponding SMPL body registrations of the interacting people. 
For the SMPL estimation task, we mainly follow the evaluation protocol of AGORA \cite{patel2021agora}. For the monocular setting, MPJPE [mm]
and MVE [mm] are calculated after alignment to the pelvis. The NMJE [mm] and
NMVE [mm] are MPJPE and MVE errors normalized by the F1 score respectively.
We adopt the Percentage of Correct Depth Relations ($\text{PCDR} ^{0.1}$ [\%]) 
metric that is introduced in \cite{sun2022bev} to evaluate depth reasoning. Contact Distances (CD [mm]) measures the distances between contact correspondences annotated in our dataset.\\
\noindent\textbf{Monocular Setting.} We evaluate one top-down method (PARE \cite{kocabas2021pare}) and two  bottom-up methods (ROMP \cite{sun2021romp} and BEV \cite{sun2022bev}) for the monocular SMPL estimation task. From \tabref{tab: smpl_estimation} we see that all methods have a relatively high MPJPE and MVE, demonstrating that current methods are not robust enough when strong human-human occlusion occurs. 
All methods fail to provide the reasonable spatial arrangement and contact relation, as shown in metric CD and \figref{fig: smpl_estimation}. \\
\textbf{Multi-View Setting.} Most of the multi-view pose estimation methods still focus only on skeleton estimation without taking body shape into account. We evaluate the open-sourced multi-view SMPL estimation method MVPose\cite{dong2019mvpose} on 4-view and 8-view settings.  
Although in the multi-view setting, MVPose achieves lower MPJPE and MVE, heavy interpenetration, and inaccurate poses in 3D space still exist, especially in the contact area (\cf \figref{fig: smpl_estimation}).  

\subsection{Detailed Geometry Reconstruction}
\label{subsec: mesh_baseline}
\noindent\textbf{Evaluation Protocol.} 
\datasetname provides high-quality 4D scans, which can serve as ground truth for the detailed geometry reconstruction task. We apply the same metrics described in \secref{subsec: mesh_eval} to measure the reconstruction accuracy.\\
\noindent\textbf{Monocular Setting.} Most of the existing monocular mesh reconstruction methods focus on the single-person scenario without any occlusion. 
We extend two methods for single-person geometry reconstruction PIFuHD \cite{saito2020pifuhd} and ICON \cite{xiu2022icon} to handle the multi-person case. For implementation details please refer to the Supp. Mat.
From \figref{fig: mesh_reconstruction}, we can observe that both methods are not robust against human-human occlusions and fail to produce high-quality reconstructions. 
\tabref{tab:mesh_reconstruction} quantitatively shows that current single-person methods cannot achieve satisfactory reconstructions when directly extended to the challenging multi-person scenario. We believe that \datasetname provides the necessary data to unlock next-generation methods to reconstruct detailed geometry from monocular RGB sequences depicting closely interacting humans.\\
\textbf{Multi-View Setting.} We evaluate the method DMC \cite{zheng2021deepmulticap} in both the 4-view and 8-view settings. 
Qualitative results in \figref{fig: mesh_reconstruction} show that although DMC can correctly reconstruct the geometry globally, artifacts still exist, \cf the hands and feet of the person colored in yellow. \tabref{tab:mesh_reconstruction} further highlights the opportunities for improvement on this task. 


\section{Conclusion}
\label{sec:coclusion}
In this paper, we propose a method to track and segment 4D scans of multiple people interacting in close range with dynamic physical contact.
To do so we first build a personalized implicit avatar model for each subject and then refine pose and shape network parameters given fused raw scans in an alternating fashion.
We further introduce \datasetname, a dataset consisting of close human interaction with high-quality 4D textured scans alongside corresponding multi-view RGB sequences, instance segmentation masks in 2D and 3D, registered parametric body models and vertex-level contact annotations. 
We define several vision benchmarks, such as monocular and multi-view human pose estimation and detailed geometry reconstruction conducted on \datasetname.

\noindent\textbf{Limitations.} 
Currently, our method does not model hands or facial expressions explicitly. We see the integration of more expressive human models \eg \cite{shen2023xavatar} as a fruitful future direction. Furthermore, the optimization schema of our method is not very computationally efficient. The optimization can be accelerated remarkably by upgrading the current deformer to a faster version \cite{chen2022fastsnarf}.

\noindent\textbf{Acknowledgments.} We thank Stefan Walter and Dean Bakker for the infrastructure support. We thank Deniz Yildiz and Laura W{\"u}lfroth for the data collection. We also thank all the participants who contribute to \datasetname.


{\small
\bibliographystyle{ieee_fullname}
\bibliography{egbib}
}

\end{document}